\documentclass[11pt]{article}
\usepackage{eacl2017}
\usepackage{times}
\usepackage{url}
\usepackage{latexsym}

\usepackage{latexsym}
\usepackage{graphicx}
\usepackage{multirow}
\usepackage{hhline}
\usepackage{stfloats}
\usepackage{setspace}
\usepackage{amsmath}
\usepackage{color}

\eaclfinalcopy 
\def\eaclpaperid{265} 

\title{Neural Tree Indexers for Text Understanding}

\author{Tsendsuren Munkhdalai \and Hong Yu\\
        University of Massachusetts, MA, USA \\ 
        \small
        {\tt \{tsendsuren.munkhdalai,hong.yu\}@umassmed.edu}
       }

\date{}

\pdfoutput=1
\begin{document}
\maketitle
\begin{abstract}
Recurrent neural networks (RNNs) process input text sequentially and model the conditional transition between word tokens. 
In contrast, the advantages of recursive networks include that they explicitly model the compositionality and the recursive structure of natural language. However, the current recursive architecture is limited by its dependence on syntactic tree. In this paper, we introduce a robust syntactic parsing-independent tree structured model, Neural Tree Indexers (NTI) that provides a middle ground between the sequential RNNs and the syntactic tree-based recursive models. NTI constructs a \textit{full n-ary tree} by processing the input text with its node function in a bottom-up fashion. Attention mechanism can then be applied to both structure and node function. We implemented and evaluated a binary-tree model of NTI, showing the model achieved the state-of-the-art performance on three different NLP tasks: natural language inference, answer sentence selection, and sentence classification, outperforming state-of-the-art recurrent and recursive neural networks \footnote{Code for the experiments and NTI is available at https://bitbucket.org/tsendeemts/nti}. 

\end{abstract}

\section{Introduction}

Recurrent neural networks (RNNs) have been successful for modeling sequence data \cite{elman:90}. RNNs equipped with gated hidden units and internal short-term memories, such as long short-term memories (LSTM) \cite{hochreiter:97} have achieved a notable success in several NLP tasks including named entity recognition \cite{lample2016neural}, constituency parsing \cite{vinyals:15a}, textual entailment recognition \cite{rocktaschel:16}, question answering \cite{hermann:15}, and machine translation \cite{bahdanau:15}. However, most LSTM models explored so far are sequential. It encodes text sequentially from left to right or vice versa and do not naturally support compositionality of language. Sequential LSTM models seem to learn syntactic structure from the natural language however their generalization on unseen text is relatively poor comparing with models that exploit syntactic tree structure \cite{bowman2015tree}. 

Unlike sequential models, recursive neural networks compose word phrases over syntactic tree structure and have shown improved performance in sentiment analysis \cite{socher2013recursive}. However its dependence on a syntactic tree architecture limits practical NLP applications.  In this study, we introduce Neural Tree Indexers (NTI), a class of tree structured models for NLP tasks. NTI takes a sequence of tokens and produces its representation by constructing a \textit{full n-ary tree} in a bottom-up fashion. Each node in NTI is associated with one of the node transformation functions: leaf node mapping and non-leaf node composition functions. Unlike previous recursive models, the tree structure for NTI is relaxed, i.e., NTI does not require the input sequences to be parsed syntactically; and therefore it is flexible and can be directly applied to a wide range of NLP tasks beyond sentence modeling. 

Furthermore, we propose different variants of node composition function and attention over tree for our NTI models. When a sequential leaf node transformer such as LSTM is chosen, the NTI network forms a sequence-tree hybrid model taking advantage of both conditional and compositional powers of sequential and recursive models. Figure $~\ref{figure:nti}$ shows a binary-tree model of NTI. Although the model does not follow the syntactic tree structure, we empirically show that it achieved the state-of-the-art performance on three different NLP applications: natural language inference, answer sentence selection, and sentence classification. 

\section{Related Work}

\begin{figure*}[t]
    \centering
        \includegraphics[width=1.0\textwidth]{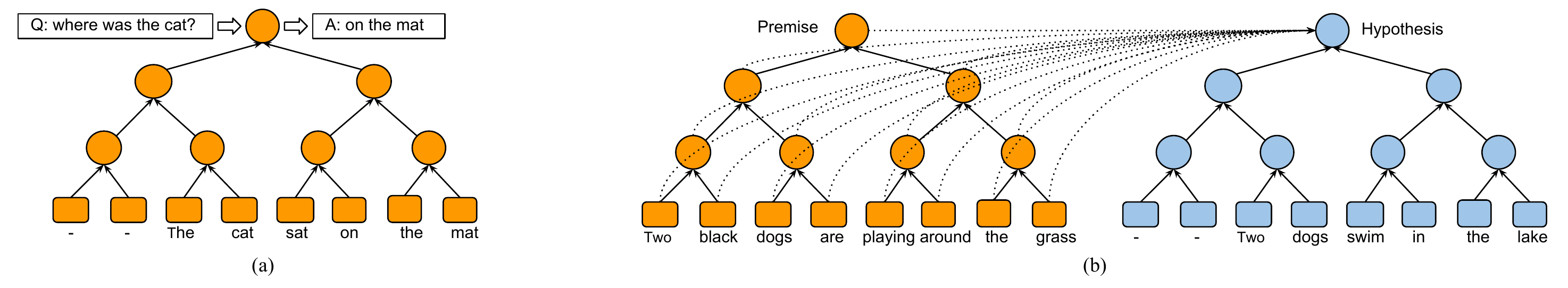}
        \caption{\label{figure:nti} A binary tree form of Neural Tree Indexers (NTI) in the context of question answering and natural language inference. We insert empty tokens (denoted by $-$) to the input text to form a full binary tree. (a) NTI produces answer representation at the root node. This representation along with the question is used to find the answer. (b) NTI learns representations for the premise and hypothesis sentences and then attentively combines them for classification. Dotted lines indicate attention over premise-indexed tree.}
        \label{figure:nti}
\end{figure*}

\subsection{Recurrent Neural Networks and Attention Mechanism}
RNNs model input text sequentially by taking a single token at each time step and producing a corresponding hidden state. The hidden state is then passed along through the next time step to provide historical sequence information. Although a great success in a variety of tasks, RNNs have limitations \cite{bengio:94,hochreiter:98}. Among them, it is not efficient at memorizing long or distant sequence \cite{sutskever:14}. This is frequently called as information flow bottleneck.
Approaches have therefore been developed to overcome the limitations. 
For example, to mitigate the information flow bottleneck, \newcite{bahdanau:15} extended RNNs with a soft attention mechanism in the context of neural machine translation, leading to improved the results in translating longer sentences. 

RNNs are linear chain-structured; this limits its potential for natural language which can be represented by complex structures including syntactic structure. In this study, we propose models to mitigate this limitation. 


\subsection{Recursive Neural Networks}

Unlike RNNs, recursive neural networks explicitly model the compositionality and the recursive structure of natural language over tree. The tree structure can be predefined by a syntactic parser \cite{socher2013recursive}. Each non-leaf tree node is associated with a node composition function which combines its children nodes and produces its own representation. The model is then trained by back-propagating error through structures \cite{goller1996learning}.

The node composition function can be varied. A single layer network with $tanh$ non-linearity was adopted in recursive auto-associate memories \cite{pollack1990recursive} and recursive autoencoders \cite{socher2011semi}. \newcite{socher2012semantic} extended this network with an additional matrix representation for each node to augment the expressive power of the model. Tensor networks have also been used as composition function for sentence-level sentiment analysis task \cite{socher2013recursive}. Recently, \newcite{zhu2015long} introduced S-LSTM which extends LSTM units to compose tree nodes in a recursive fashion. 

In this paper, we introduce a novel attentive node composition function that is based on S-LSTM. Our NTI model does not rely on either a parser output or a fine-grained supervision of non-leaf nodes, both required in previous work. In NTI, the supervision from the target labels is provided at the root node. As such, our NTI model is robust and applicable to a wide range of NLP tasks. We introduce attention over tree in NTI to overcome the vanishing/explode gradients challenges as shown in RNNs. 

\section{Methods}

Our training set consists of $N$ examples $\lbrace X^i,Y^i \rbrace^N_{i=1}$, where the input $X^i$ is a sequence of word tokens $w^i_1, w^i_2, \ldots, w^i_{T_i}$ and the output $Y^i$ can be either a single target or a sequence. Each input word token $w_t$ is represented by its word embedding $x_t \in R^k$.

NTI is a \textit{full n-ary tree} (and the sub-trees can be overlapped). It has two types of transformation function: non-leaf node function $f^{node}(h^1, \ldots, h^c)$ and leaf node function $f^{leaf}(x_t)$. $f^{leaf}(x_t)$ computes a (possibly non-linear) transformation of the input word embedding $x_t$. $f^{node}(h^1, \ldots, h^c)$ is a function of its child nodes representation $h^1, \ldots, h^c$, where  $c$ is the total number of child nodes of this non-leaf node. 

NTI can be implemented with different tree structures. In this study we implemented and evaluated a binary tree form of NTI: a non-leaf node can take in only two direct child nodes (i.e., $c=2$). Therefore, the function $f^{node}(h^l, h^r)$ composes its left child node $h^l$ and right child node $h^r$. Figure $~\ref{figure:nti}$ illustrates our NTI model that is applied to question answering (a) and natural language inference tasks (b). Note that the node and leaf node functions are neural networks and are the only training parameters in NTI.  

We explored two different approaches to compose node representations: an extended LSTM and attentive node composition functions, to be described below.    

\subsection{Non-Leaf Node Composition Functions}
We define two different methods for non-leaf node function $f^{node}(h^l, h^r)$.

\textbf{LSTM-based Non-leaf Node Function (S-LSTM):} We initiate $f^{node}(h^l, h^r)$ with LSTM. For non-leaf node, we adopt S-LSTM \newcite{zhu2015long}, an extension of LSTM to tree structures, to learn a node representation by its children nodes. Let $h^l_t$, $h^r_t$, $c^l_t$ and $c^r_t$ be vector representations and cell states for the left and right children. An S-LSTM computes a parent node representation $h^p_{t+1}$ and a node cell state $c^p_{t+1}$ as
\begin{equation}
i_{t+1}=\sigma ( W^{s}_1 h^l_t+W^{s}_2 h^r_t + W^{s}_3 c^l_t + W^{s}_4 c^r_t)
\end{equation}
\begin{equation}
f^l_{t+1}=\sigma ( W^{s}_5 h^l_t+W^{s}_6 h^r_t + W^{s}_7 c^l_t + W^{s}_8 c^r_t)
\end{equation}
\begin{equation}
f^r_{t+1}=\sigma ( W^{s}_9 h^l_t+W^{s}_{10} h^r_t + W^{s}_{11} c^l_t + W^{s}_{12} c^r_t)
\end{equation}
\begin{multline}
c^p_{t+1}= f^l_{t+1} \odot c^l_t + f^r_{t+1} \odot c^r_t \\ + i_{t+1} \odot \tanh(W^{s}_{13} h^l_t + W^{s}_{14} h^r_t)
\end{multline}
\begin{equation}
o_{t+1}=\sigma ( W^{s}_{15} h^l_t+W^{s}_{16} h^r_t + W^{s}_{18} c^p_{t+1})
\end{equation}
\begin{equation}
h^p_{t+1}=o_{t+1} \odot tanh(c^p_{t+1})
\end{equation}
where $W^{s}_1, \ldots, W^{s}_{18} \in R^{k \times k}$ and biases (for brevity we eliminated the bias terms) are the training parameters.  $\sigma$ and $\odot$ denote the element-wise $sigmoid$ function and the element-wise vector multiplication. Extension of S-LSTM non-leaf node function to compose more children is straightforward. However, the number of parameters increases quadratically in S-LSTM as we add more child nodes.

\textbf{Attentive Non-leaf Node Function (ANF):} Some NLP applications (e.g., QA and machine translation) would benefit from a dynamic query dependent composition function. We introduce ANF as a new non-leaf node function. Unlike S-LSTM, ANF composes the child nodes attentively in respect to another relevant input vector $q \in R^{k}$. The input vector $q$ can be a learnable representation from a sequence representation.
Given a matrix $S^{ANF} \in R^{k \times 2}$ resulted by concatenating the child node representations $h^l_t$, $h^r_t$ and the third input vector $q$, ANF is defined as
\begin{equation}
m=f^{score} (S^{ANF}, q)
\end{equation}
\begin{equation}
\alpha=softmax(m)
\end{equation}
\begin{equation}
z=S^{ANF} \alpha ^{\top}
\end{equation}
\begin{equation}
h^p_{t+1}=ReLU (W^{ANF}_1 z)
\end{equation}
where $W^{ANF}_1 \in R^{k \times k}$ is a learnable matrix, $m \in R^2$ the attention score and $\alpha \in R^2$ the attention weight vector for each child. $f^{score}$ is an attention scoring function, which can be implemented as a multi-layer perceptron (MLP)
\begin{multline}
m=w^{\top} ReLU (W^{score}_1 S^{ANF}  \\ + W^{score}_2 q \otimes e)
\end{multline}
or a matrix-vector product $m = q^{\top} S^{ANF} $. The matrices $W^{score}_1$ and $ W^{score}_2 \in R^{k \times k}$ and the vector $w \in R^{k}$ are training parameters. $e \in R^2$ is a vector of ones and $\otimes$ the outer product. We use $ReLU$ function for non-linear transformation.

\subsection{Attention Over Tree}

\begin{table*}[t]
\begin{center}
\small
\begin{tabular}{c|c|c|c|c}
\hline 
Model & $d$ & $|\theta|_M$ & Train	& Test \\
\hline
\multicolumn{1}{l|}{Classifier with handcrafted features \cite{bowman:15}} & \multicolumn{1}{|r|}{-} & \multicolumn{1}{|r|}{-} & \multicolumn{1}{|r|}{99.7} & \multicolumn{1}{|r}{78.2} \\
\hline
\multicolumn{1}{l|}{LSTMs encoders \cite{bowman:15}} & \multicolumn{1}{|r|}{300} & \multicolumn{1}{|r|}{3.0M} & \multicolumn{1}{|r|}{83.9} & \multicolumn{1}{|r}{80.6} \\
\multicolumn{1}{l|}{Dependency Tree CNN encoders \cite{Lili16}} & \multicolumn{1}{|r|}{300} & \multicolumn{1}{|r|}{3.5M} & \multicolumn{1}{|r|}{83.3} & \multicolumn{1}{|r}{82.1} \\
\multicolumn{1}{l|}{NTI-SLSTM (Ours)} & \multicolumn{1}{|r|}{300} & \multicolumn{1}{|r|}{3.3M} & \multicolumn{1}{|r|}{83.9} & \multicolumn{1}{|r}{82.4} \\
\multicolumn{1}{l|}{SPINN-PI encoders \cite{BowmanGRGMP16}} & \multicolumn{1}{|r|}{300} & \multicolumn{1}{|r|}{3.7M} & \multicolumn{1}{|r|}{89.2} & \multicolumn{1}{|r}{83.2} \\
\multicolumn{1}{l|}{NTI-SLSTM-LSTM (Ours)} & \multicolumn{1}{|r|}{300} & \multicolumn{1}{|r|}{4.0M} & \multicolumn{1}{|r|}{82.5} & \multicolumn{1}{|r}{83.4} \\ 
\hline
\multicolumn{1}{l|}{LSTMs attention \cite{rocktaschel:16}} & \multicolumn{1}{|r|}{100} & \multicolumn{1}{|r|}{242K} & \multicolumn{1}{|r|}{85.4} & \multicolumn{1}{|r}{82.3} \\
\multicolumn{1}{l|}{LSTMs word-by-word attention \cite{rocktaschel:16}} & \multicolumn{1}{|r|}{100} & \multicolumn{1}{|r|}{250K} & \multicolumn{1}{|r|}{85.3} & \multicolumn{1}{|r}{83.5} \\
\multicolumn{1}{l|}{NTI-SLSTM node-by-node global attention (Ours)} & \multicolumn{1}{|r|}{300} & \multicolumn{1}{|r|}{3.5M} & \multicolumn{1}{|r|}{85.0} & \multicolumn{1}{|r}{84.2} \\
\multicolumn{1}{l|}{NTI-SLSTM node-by-node tree attention (Ours)} & \multicolumn{1}{|r|}{300} & \multicolumn{1}{|r|}{3.5M} & \multicolumn{1}{|r|}{86.0} & \multicolumn{1}{|r}{84.3} \\
\multicolumn{1}{l|}{NTI-SLSTM-LSTM node-by-node tree attention (Ours)} & \multicolumn{1}{|r|}{300} & \multicolumn{1}{|r|}{4.2M} & \multicolumn{1}{|r|}{88.1} & \multicolumn{1}{|r}{85.7} \\
\multicolumn{1}{l|}{NTI-SLSTM-LSTM node-by-node global attention (Ours)} & \multicolumn{1}{|r|}{300} & \multicolumn{1}{|r|}{4.2M} & \multicolumn{1}{|r|}{87.6} & \multicolumn{1}{|r}{85.9} \\
\multicolumn{1}{l|}{mLSTM word-by-word attention \cite{WangJ15b}} & \multicolumn{1}{|r|}{300} & \multicolumn{1}{|r|}{1.9M} & \multicolumn{1}{|r|}{92.0} & \multicolumn{1}{|r}{86.1} \\
\multicolumn{1}{l|}{LSTMN with deep attention fusion \cite{ChengDL16}} & \multicolumn{1}{|r|}{450} & \multicolumn{1}{|r|}{3.4M} & \multicolumn{1}{|r|}{88.5} & \multicolumn{1}{|r}{86.3} \\
\multicolumn{1}{l|}{Tree matching NTI-SLSTM-LSTM tree attention (Ours)} & \multicolumn{1}{|r|}{300} & \multicolumn{1}{|r|}{3.2M} & \multicolumn{1}{|r|}{87.3} & \multicolumn{1}{|r}{86.4} \\
\multicolumn{1}{l|}{Decomposable Attention Model \cite{parikh2016decomposable}} & \multicolumn{1}{|r|}{200} & \multicolumn{1}{|r|}{580K} & \multicolumn{1}{|r|}{90.5} & \multicolumn{1}{|r}{86.8} \\
\multicolumn{1}{l|}{Tree matching NTI-SLSTM-LSTM global attention (Ours)} & \multicolumn{1}{|r|}{300} & \multicolumn{1}{|r|}{3.2M} & \multicolumn{1}{|r|}{87.6} & \multicolumn{1}{|r}{87.1} \\
\multicolumn{1}{l|}{Full tree matching NTI-SLSTM-LSTM global attention (Ours)} & \multicolumn{1}{|r|}{300} & \multicolumn{1}{|r|}{3.2M} & \multicolumn{1}{|r|}{88.5} & \multicolumn{1}{|r}{\textbf{87.3}} \\
\hline
\end{tabular}
\end{center}
\caption{\label{table:snli}Training and test accuracy on natural language inference task. $d$ is the word embedding size and $|\theta|_M$ the number of model parameters.}
\end{table*}

Comparing with sequential LSTM models, NTI has less recurrence, which is defined by the tree depth, $log(n)$ for binary tree where $n$ is the length of the input sequence. However, NTI still needs to compress all the input information into a single representation vector of the root. This imposes practical difficulties when processing long sequences. We address this issue with attention mechanism over tree. In addition, the attention mechanism can be used for matching trees (described in Section 4 as Tree matching NTI) that carry different sequence information. We first define a global attention and then introduce a tree attention which considers the parent-child dependency for calculation of the attention weights.

\textbf{Global Attention:} An attention neural network for the global attention takes all node representations as input and produces an attentively blended vector for the whole tree. This neural net is similar to ANF. Particularly, given a matrix $S^{GA} \in R^{k \times 2n-1}$ resulted by concatenating the node representations $h_1$, \ldots, $h_{2n-1}$ and the relevant input representation $q$, the global attention is defined as 
\begin{equation}
m=f^{score} (S^{GA}, q)
\end{equation}
\begin{equation}
\alpha=softmax(m)
\end{equation}
\begin{equation}
z=S^{GA} \alpha ^{\top}
\end{equation}
\begin{equation}
h^{tree}=ReLU (W^{GA}_1 z + W^{GA}_2 q)
\end{equation}
where $W^{GA}_1$ and $W^{GA}_2 \in R^{k \times k}$ are training parameters and $\alpha \in R^{2n-1}$ the attention weight vector for each node. This attention mechanism is robust as it globally normalizes the attention score $m$ with $softmax$ to obtain the weights $\alpha$. However, it does not consider the tree structure when producing the final representation $h^{tree}$.

\textbf{Tree Attention:} We modify the global attention network to the tree attention mechanism. The resulting tree attention network performs almost the same computation as ANF for each node. It compares the parent and children nodes to produce a new representation assuming that all node representations are constructed. Given a matrix $S^{TA} \in R^{k \times 3}$ resulted by concatenating the parent node representation $h^p_t$, the left child $h^l_t$ and the right child $h^r_t$ and the relevant input representation $q$, every non-leaf node $h^p_t$ simply updates its own representation by using the following equation in a bottom-up manner.
\begin{equation}
m=f^{score} (S^{TA}, q)
\end{equation}
\begin{equation}
\alpha=softmax(m)
\end{equation}
\begin{equation}
z=S^{TA} \alpha ^{\top}
\end{equation}
\begin{equation}
h^{p}_{t}=ReLU (W^{TA}_1 z)
\end{equation}
and this equation is similarity to the global attention. However, now each non-leaf node attentively collects its own and children representations and passes towards the root which finally constructs the attentively blended tree representation. Note that unlike the global attention, the tree attention locally normalizes the attention scores with $softmax$.

\section{Experiments}

We describe in this section experiments on three different NLP tasks, natural language inference, question answering and sentence classification to demonstrate the flexibility and the effectiveness of NTI in the different settings.

We trained NTI using Adam \cite{kingma:14} with hyperparameters selected on development set. The pre-trained 300-D Glove 840B vectors \cite{pennington:14} were obtained for the word embeddings\footnote{http://nlp.stanford.edu/projects/glove/}. The word embeddings are fixed during training. The embeddings for out-of-vocabulary words were set to zero vector. We pad the input sequence to form a \textit{full binary tree}. A padding vector was inserted when padding. We analyzed the effects of the padding size and found out that it has no influence on the performance (see Appendix \ref{sup:pad}). The size of hidden units of the NTI modules were set to 300. The models were regularized by using dropouts and an $l_2$ weight decay.\footnote{More detail on hyper-parameters can be found in code.}

\subsection{Natural Language Inference}

We conducted experiments on the Stanford Natural Language Inference (SNLI) dataset \cite{bowman:15}, which consists of 549,367/9,842/9,824 premise-hypothesis pairs for train/dev/test sets and target label indicating their relation. Unless otherwise noted, we follow the setting in the previous work \cite{Lili16,BowmanGRGMP16} and use an MLP 
for classification which takes in NTI outputs and computes the concatenation $[h^p_{2n-1};h^h_{2n-1}]$, absolute difference $h^p_{2n-1}-h^h_{2n-1}$ and elementwise product $h^p_{2n-1} \cdot h^h_{2n-1}$ of the two sentence representations. The MLP has also an input layer with 1024 units with $ReLU$ activation and a $softmax$ output layer. We explored nine different task-oriented NTI models with varying complexity, to be described below. For each model, we set the batch size to 32. The initial learning, the regularization strength and the number of epoch to be trained are varied for each model.

\textbf{NTI-SLSTM:} this model does not rely on $f^{leaf}$ transformer but uses the S-LSTM units for the non-leaf node function. We set the initial learning rate to 1e-3 and $l_2$ regularizer strength to 3e-5, and train the model for 90 epochs. The neural net was regularized by 10\% input dropouts and the 20\% output dropouts. 

\textbf{NTI-SLSTM-LSTM:} we use LSTM for the leaf node function $f^{leaf}$. Concretely, the LSTM output vectors are given to NTI-SLSTM and the memory cells of the lowest level S-LSTM were initialized with the LSTM memory states. The hyper-parameters are the same as the previous model.



\textbf{NTI-SLSTM node-by-node global attention:} This model learns inter-sentence relation with the global attention over premise-indexed tree, which is similar to word-by-word attention model of \newcite{rocktaschel:16} in that it attends over the premise tree nodes at every time step of hypothesis encoding. We tie the weight parameters of the two NTI-SLSTMs for premise and hypothesis and no $f^{leaf}$ transformer used. We set the initial learning rate to 3e-4 and $l_2$ regularizer strength to 1e-5, and train the model for 40 epochs. The neural net was regularized by 15\% input dropouts and the 15\% output dropouts.

\textbf{NTI-SLSTM node-by-node tree attention:} this is a variation of the previous model with the tree attention. The hyper-parameters are the same as the previous model.

\textbf{NTI-SLSTM-LSTM node-by-node global attention:} in this model we include LSTM as the leaf node function $f^{leaf}$. Here we initialize the memory cell of S-LSTM with LSTM memory and hidden/memory state of hypothesis LSTM with premise LSTM (the later follows the work of \cite{rocktaschel:16}). We set the initial learning rate to 3e-4 and $l_2$ regularizer strength to 1e-5, and train the model for 10 epochs. The neural net was regularized by 10\% input dropouts and the 15\% output dropouts.

\textbf{NTI-SLSTM-LSTM node-by-node tree attention:} this is a variation of the previous model with the tree attention. The hyper-parameters are the same as the previous model.

\textbf{Tree matching NTI-SLSTM-LSTM global attention:} this model first constructs the premise and hypothesis trees simultaneously with the NTI-SLSTM-LSTM model and then computes their matching vector by using the global attention and an additional LSTM. The attention vectors are produced at each hypothesis tree node and then are given to the LSTM model sequentially. The LSTM model compress the attention vectors and outputs a single matching vector, which is passed to an MLP for classification. The MLP for this tree matching setting has an input layer with 1024 units with $ReLU$ activation and a $softmax$ output layer.

Unlike \newcite{WangJ15b}'s matching LSTM model which is specific to matching sequences, 
we use the standard LSTM units and match trees.
We set the initial learning rate to 3e-4 and $l_2$ regularizer strength to 3e-5, and train the model for 20 epochs. The neural net was regularized by 20\% input dropouts and the 20\% output dropouts.

\textbf{Tree matching NTI-SLSTM-LSTM tree attention:} we replace the global attention with the tree attention. The hyper-parameters are the same as the previous model.

\textbf{Full tree matching NTI-SLSTM-LSTM global attention:} this model produces two sets of the attention vectors, one by attending over the premise tree regarding each hypothesis tree node and another by attending over the hypothesis tree regarding each premise tree node. Each set of the attention vectors is given to a LSTM model to achieve full tree matching. The last hidden states of the two LSTM models (i.e. one for each attention vector set) are concatenated for classification. The training weights are shared among the LSTM models The hyper-parameters are the same as the previous model.\footnote{Computational constraint prevented us from experimenting the tree attention variant of this model}

Table \ref{table:snli} shows the results of our models. For comparison, we include the results from the published state-of-the-art systems.
While most of the sentence encoder models rely solely on word embeddings, the dependency tree CNN and the SPINN-PI models make use of sentence parser output; which present strong baseline systems. The last set of methods designs inter-sentence relation with soft attention \cite{bahdanau:15}. Our best score on this task is 87.3\% accuracy obtained with the full tree matching NTI model. The previous best performing model on the task performs phrase matching by using the attention mechanism. 

Our results show that NTI-SLSTM improved the performance of the sequential LSTM encoder by approximately 2\%. Not surprisingly, using LSTM as leaf node function helps in learning better representations. Our NTI-SLSTM-LSTM is a hybrid model which encodes a sequence sequentially through its leaf node function and then hierarchically composes the output representations. The node-by-node attention models improve the performance, indicating that modeling inter-sentence interaction is an important element in NLI. Aggregating matching vector between trees or sequences with a separate LSTM model is effective. The global attention seems to be robust on this task. The tree attention were not helpful as it normalizes the attention scores locally in parent-child relationship.

\subsection{Answer Sentence Selection}


\begin{table}[t]
\begin{center}
\small
\begin{tabular}{c|c|c}
\hline 
Model & MAP & MRR \\
\hline
\multicolumn{1}{l|}{Classifier with features \shortcite{yih13}} & \multicolumn{1}{|r|}{0.5993} & \multicolumn{1}{|r}{0.6068} \\
\hline
\multicolumn{1}{l|}{Paragraph Vector \shortcite{le2014}} & \multicolumn{1}{|r|}{0.5110} & \multicolumn{1}{|r}{0.5160} \\
\multicolumn{1}{l|}{Bigram-CNN \shortcite{yu2014}} & \multicolumn{1}{|r|}{0.6190} & \multicolumn{1}{|r}{0.6281} \\
\multicolumn{1}{l|}{3-layer LSTM \shortcite{miao2016}} & \multicolumn{1}{|r|}{0.6552} & \multicolumn{1}{|r}{0.6747} \\
\multicolumn{1}{l|}{3-layer LSTM attention \shortcite{miao2016}} & \multicolumn{1}{|r|}{0.6639} & \multicolumn{1}{|r}{0.6828} \\
\multicolumn{1}{l|}{NASM \shortcite{miao2016}} & \multicolumn{1}{|r|}{0.6705} & \multicolumn{1}{|r}{\bf 0.6914} \\
\multicolumn{1}{l|}{NTI (Ours)} & \multicolumn{1}{|r|}{\bf 0.6742} & \multicolumn{1}{|r}{0.6884} \\
\hline
\end{tabular}
\end{center}
\caption{\label{table:qa}Test set performance on answer sentence selection.}
\end{table}

\begin{table}[t]
\begin{center}
\small
\begin{tabular}{c|c|c}
\hline 
Model & Bin & FG \\
\hline
\multicolumn{1}{l|}{RNTN \cite{socher2013recursive}} & \multicolumn{1}{|r|}{85.4} & \multicolumn{1}{|r}{45.7} \\
\multicolumn{1}{l|}{CNN-MC \cite{kim:13}} & \multicolumn{1}{|r|}{88.1} & \multicolumn{1}{|r}{47.4} \\
\multicolumn{1}{l|}{DRNN \cite{irsoy15}} & \multicolumn{1}{|r|}{86.6} & \multicolumn{1}{|r}{49.8} \\
\multicolumn{1}{l|}{2-layer LSTM \cite{tai2015improved}} & \multicolumn{1}{|r|}{86.3} & \multicolumn{1}{|r}{46.0} \\
\multicolumn{1}{l|}{Bi-LSTM \cite{tai2015improved}} & \multicolumn{1}{|r|}{87.5} & \multicolumn{1}{|r}{49.1} \\
\multicolumn{1}{l|}{NTI-SLSTM (Ours)} & \multicolumn{1}{|r|}{87.8} & \multicolumn{1}{|r}{50.5} \\
\multicolumn{1}{l|}{CT-LSTM \cite{tai2015improved}} & \multicolumn{1}{|r|}{88.0} & \multicolumn{1}{|r}{51.0} \\
\multicolumn{1}{l|}{DMN \cite{ankit16}} & \multicolumn{1}{|r|}{88.6} & \multicolumn{1}{|r}{52.1} \\
\multicolumn{1}{l|}{NTI-SLSTM-LSTM (Ours)} & \multicolumn{1}{|r|}{\bf  89.3} & \multicolumn{1}{|r}{\bf 53.1} \\
\hline
\end{tabular}
\end{center}
\caption{\label{table:sent}Test accuracy for sentence classification. Bin: binary, FG: fine-grained 5 classes.}
\end{table}

\begin{figure*}[t]
\hspace*{-0.8cm}
    \centering
        \includegraphics[width=1.0\textwidth]{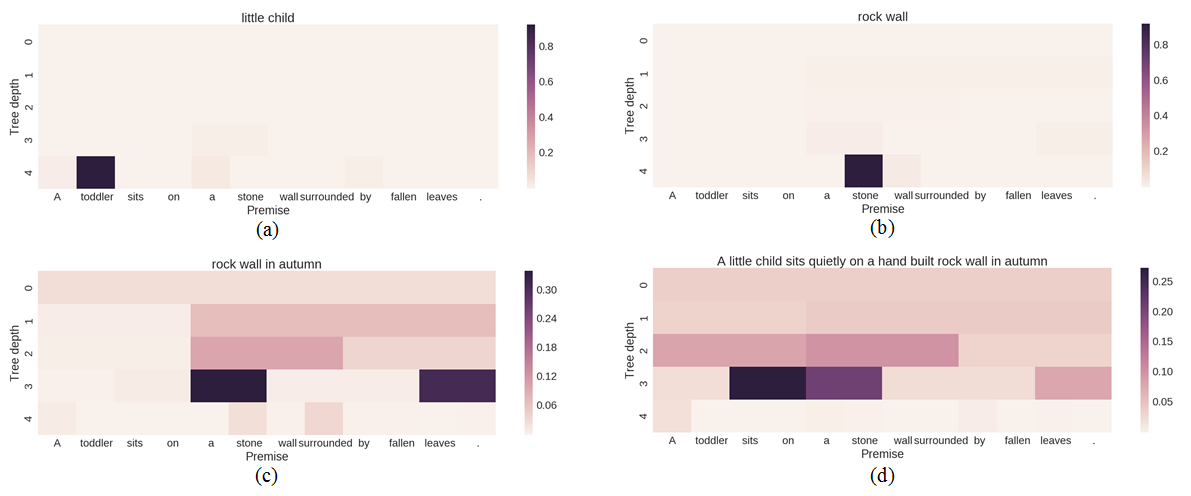}
        \caption{Node-by-node attention visualizations. The phrases shown on the top are nodes from hypothesis-indexed tree and the premise tokens are listed along the x-axis. The adjacent cells are composed in the top cell representing a binary tree and resulting a longer attention span.}
        \label{figure:at}
\end{figure*}

For this task, a model is trained to identify the correct sentences that answer a factual question, from a set of candidate sentences. We experiment on WikiQA dataset constructed from Wikipedia \cite{yang2015wikiqa}. The dataset contains 20,360/2,733/6,165 QA pairs for train/dev/test sets.

We used the same setup in the language inference task except that we replace the $softmax$ layer with a $sigmoid$ layer and model the following conditional probability distribution.
\begin{equation}
p_{\theta}(y=1|h^q_n,h^a_n) = sigmoid(o^{QA})
\end{equation}
where $h^q_n$ and $h^a_n$ are the question and the answer encoded vectors and $o^{QA}$ denotes the output of the hidden layer of the MLP. 
For this task, we use NTI-SLSTM-LSTM to encode answer candidate sentences and NTI-ANF-LSTM to encode the question sentences. Note that NTI-ANF-LSTM is relied on ANF as the non-leaf node function. $q$ vector for NTI-ANF-LSTM is the answer representation produced by the answer encoding NTI-SLSTM-LSTM model. We set the batch size to 4 and the initial learning rate to 1e-3, and train the model for 10 epochs. We used 20\% input dropouts and no $l_2$ weight decay. Following previous work, we adopt MAP and MRR as the evaluation metrics for this task.\footnote{We used \textit{trec\_eval} script to calculate the evaluation metrics}

Table \ref{table:qa} presents the results of our model and the previous models for the task.\footnote{Inclusion of simple word count feature improves the performance by around 0.15-0.3 across the board} The classifier with handcrafted features is a SVM model trained with a set of features. The Bigram-CNN model is a simple convolutional neural net. The Deep LSTM and LSTM attention models outperform the previous best result by a large margin, nearly 5-6\%. NASM improves the result further and sets a strong baseline by combining variational auto-encoder \cite{kingma2014auto} with the soft attention. In NASM, they adopt a deep three-layer LSTM and introduced a latent stochastic attention mechanism over the answer sentence. Our NTI model exceeds NASM by approximately 0.4\% on MAP for this task.

\begin{table*}[t]
\begin{center}
\hspace*{-0.6cm}
\small
\begin{tabular}{l|l|l|l}
\hline 
\bf a person & \bf park for fun & \bf Santa Claus & \bf sad, depressed, and hatred  \\
\hline
\multicolumn{1}{l|}{single person} & \multicolumn{1}{|l|}{an outdoor concert at the park} & \multicolumn{1}{|l|}{a snowmobile in a blizzard} & \multicolumn{1}{|l}{an Obama supporter is upset} \\
\multicolumn{1}{l|}{a woman} & \multicolumn{1}{|l|}{kids playing at a park outside} & \multicolumn{1}{|l|}{a Skier ski - jumping } & \multicolumn{1}{|l}{but doesn't have any money} \\
\multicolumn{1}{l|}{a young person} & \multicolumn{1}{|l|}{a mom takes a break in a park} & \multicolumn{1}{|l|}{A skier preparing a trick} & \multicolumn{1}{|l}{crying because he didn't get cake} \\
\multicolumn{1}{l|}{a guy} & \multicolumn{1}{|l|}{people play frisbee outdoors} & \multicolumn{1}{|l|}{a child is playing on christmas} & \multicolumn{1}{|l}{trying his hardest to not fall off} \\
\multicolumn{1}{l|}{a single human} & \multicolumn{1}{|l|}{takes his lunch break in the park} & \multicolumn{1}{|l|}{two men play with a snowman} & \multicolumn{1}{|l}{is upset and crying on the ground} \\
\hline
\end{tabular}
\end{center}
\caption{\label{table:phrase}Nearest-neighbor phrases based on cosine similarity between learned representations.}
\end{table*}

\subsection{Sentence Classification}

Lastly, we evaluated NTI on the Stanford Sentiment Treebank (SST) \cite{socher2013recursive}. This dataset comes with standard train/dev/test sets and two subtasks: binary sentence classification or fine-grained classification of five classes. We trained our model on the text spans corresponding to labeled phrases in the training set and evaluated the model on the full sentences.

We use NTI-SLSTM and NTI-SLSTM-LSTM models to learn sentence representations for the task. The sentence representations were passed to a two-layer MLP for classification. 
We set the batch size to 64, the initial learning rate to 1e-3 and $l_2$ regularizer strength to 3e-5, and train each model for 10 epochs. The NTI-SLSTM model was regularized by 10\%/20\% of input/output and 20\%/30\% of input/output dropouts and the NTI-SLSTM-LSTM model 20\% of input and 20\%/30\% of input/output dropouts for binary and fine-grained settings.

NTI-SLSTM-LSTM (as shown in Table \ref{table:sent}) set the state-of-the-art results on both subtasks.
Our NTI-SLSTM model performed slightly worse than its constituency tree-based counter part, CT-LSTM model. The CT-LSTM model composes phrases according to the output of a sentence parser and uses a node composition function similar to S-LSTM. After we transformed the input with the LSTM leaf node function, we achieved the best performance on this task. 

\begin{table*}[t]
\begin{center}
\small
\begin{tabular}{l|l|l}
\hline 
\bf A dog mouth holds a retrieved ball. & \bf A cat nurses puppies. & \bf \bf A dog sells a woman a hat. \\
\hline
\multicolumn{1}{p{5cm}|}{A brown and white dog holds  a tennis ball in his mouth.} & \multicolumn{1}{|p{5cm}|}{A golden retriever nurses some other dogs puppies.} & \multicolumn{1}{|l}{The dog is a labrador retriever.} 
\\
\multicolumn{1}{l|}{The dog has a ball.} & \multicolumn{1}{|l|}{A golden retriever nurses puppies.} & \multicolumn{1}{|l}{A girl is petting her dog.} 
\\
\multicolumn{1}{l|}{The dogs are chasing a ball.} & \multicolumn{1}{|p{5cm}|}{A mother dog checking up on her baby puppy.} & \multicolumn{1}{|l}{The dog is a shitzu.} 
\\
\multicolumn{1}{l|}{A small dog runs to catch a ball.} & \multicolumn{1}{|l|}{A girl is petting her dog.} & \multicolumn{1}{|l}{A husband and wife making pizza.} 
\\
\multicolumn{1}{l|}{The puppy is chasing a ball.} & \multicolumn{1}{|l|}{The hat wearing girl is petting a cat.} & \multicolumn{1}{|l}{The dog is a chihuahua.} \\
\hline
\end{tabular}
\end{center}
\caption{\label{table:sent}Nearest-neighbor sentences based on cosine similarity between learned representations.}
\end{table*}

\begin{figure}[b]
\vspace*{-0.7cm}
\hspace*{-1.2cm}
    \centering
        \includegraphics[width=0.6\textwidth]{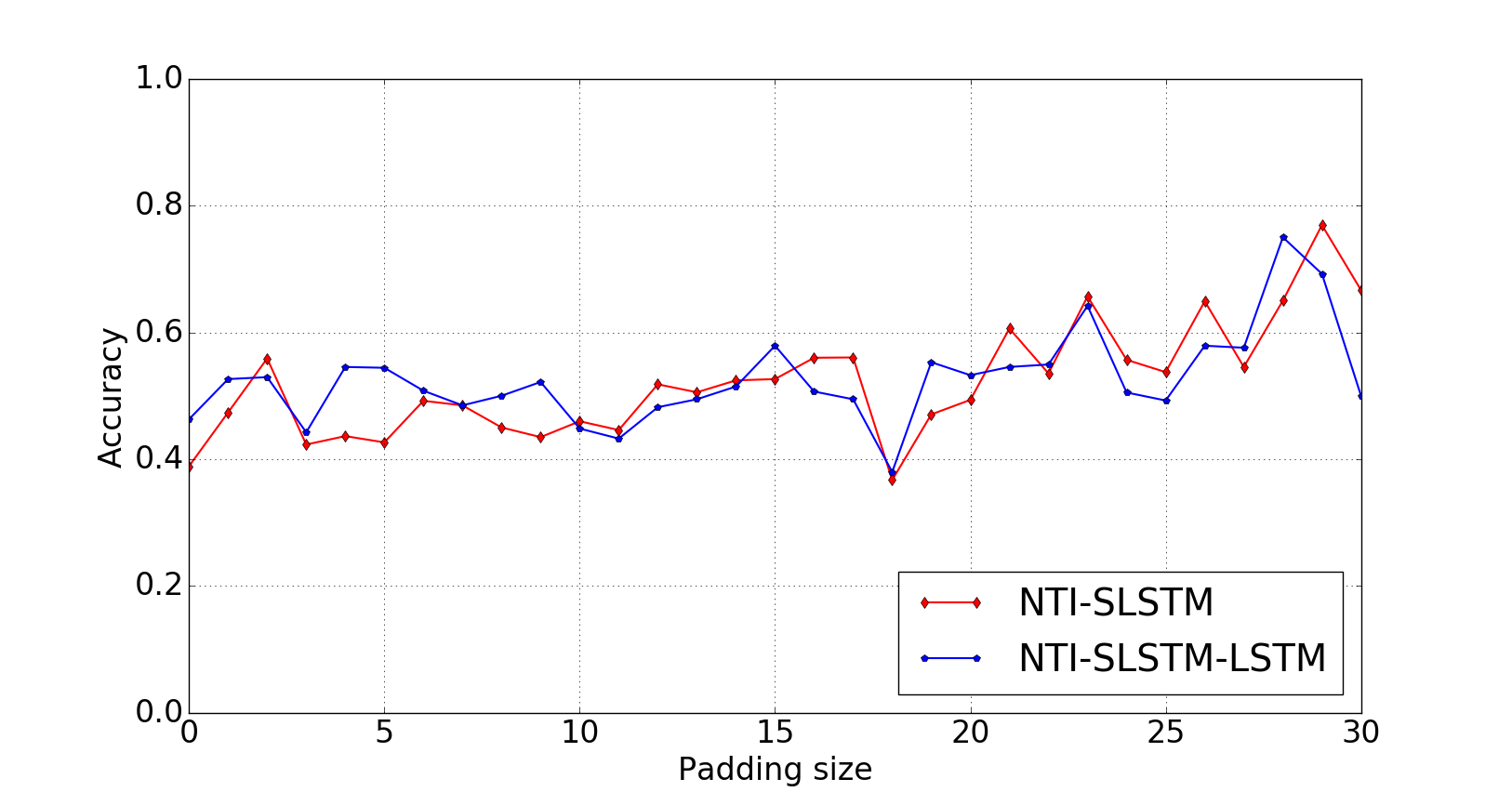}
        \caption{Fine-grained sentiment classification accuracy vs. padding size on test set of SST data.}
        \label{figure:pad}
\end{figure}

\section{Qualitative Analysis}

\subsection{Attention and Compositionality}
To help analyzing the results, we output attention weights by our NTI-SLSTM node-by-node global attention model. Figure \ref{figure:at} shows the attention heatmaps for two sentences in the SNLI test set. It shows that our model semantically aligns single or multiword expressions (\textit{"little child"} and \textit{"toddler"}; \textit{"rock wall"} and \textit{"stone"}). 
In addition, our model is able to re-orient its attention over different parts of the hypothesis when the expression is more complex. For example, for (c) \textit{"rock wall in autumn"}, NTI mostly focuses on the nodes in depth 1, 2 and 3 representing contexts related to \textit{"a stone"}, \textit{"leaves."} and \textit{"a stone wall surrounded"}. Surprisingly, attention degree for the single word expression like \textit{"stone"}, \textit{"wall"} and \textit{"leaves"} is lower to compare with multiword phrases. Sequence models lack this property as they have no explicit composition module to produce such mutiword phrases. 

Finally, the most interesting pattern is that the model attends over higher level (low depth) tree nodes with rich semantics when considering a (c) longer phrase or (d) full sentence. As shown in (d), the NTI model aligns the root node representing the whole hypothesis sentence to the higher level tree nodes covering larger sub-trees in the premise. It certainly ignores the lower level single word expressions and only starts to attend when the words are collectively to form rich semantics.

\subsection{Learned Representations of Phrases and Sentences}

Using cosine similarity 
between their representations produced by the NTI-SLSTM model, we show that NTI is able to capture paraphrases on SNLI test data. As shown in Table \ref{table:phrase}, NTI seems to distinguish plural from singular forms (similar phrases to \textit{"a person"}).  
In addition, NTI captures non-surface knowledge. For example, the phrases similar to \textit{"park for fun"} tend to align to the semantic content of \textit{fun} and \textit{park}, including \textit{"people play frisbee outdoors"}. The NTI model was able to relate \textit{"Santa Claus"} to \textit{christmas} and \textit{snow}. Interestingly, the learned representations were also able to connect implicit semantics. For example, NTI found that \textit{"sad, depressed, and hatred"} is close to the phrases like \textit{"an Obama supporter is upset"}. Overall the NTI model is robust to the length of the phrases being matched. Given a short phrase, NTI can retrieve longer yet semantically coherent sequences from the SNLI test set. 

In Table \ref{table:sent}, we show nearest-neighbor sentences from SNLI test set. Note that the sentences listed in the first two columns sound semantically coherent but not the ones in the last column. The query sentence \textit{"A dog sells a women a hat"} does not actually represent a common-sense knowledge and this sentence now seem to confuse the NTI model. As a result, the retrieved sentence are arbitrary and not coherent.

\subsection{Effects of Padding Size}
\label{sup:pad}

We introduced a special padding character in order to construct full binary tree. Does this padding character influence the performance of the NTI models? In Figure \ref{figure:pad}, we show relationship between the padding size and the accuracy on Stanford sentiment analysis data. Each sentence was padded to form a full binary tree. The x-axis represents the number of padding characters introduced. When the padding size is less (up to 10), the NTI-SLSTM-LSTM model performs better. However, this model tends to perform poorly or equally when the padding size is large. Overall we do not observe any significant performance drop for both models as the padding size increases. This suggests that NTI learns to ignore the special padding character while processing padded sentences. The same scenario was also observed while analyzing attention weights. The attention over the padded nodes was nearly zero.

\section{Discussion and Conclusion}
We introduced Neural Tree Indexers, a class of tree structured recursive neural network. The NTI models achieved state-of-the-art performance on different NLP tasks. 
Most of the NTI models form deep neural networks and we think this is one reason that NTI works well even if it lacks direct linguistic motivations followed by other syntactic-tree-structured recursive models \cite{socher2013recursive}. 

CNN and NTI are topologically related \cite{nal2013}. Both NTI and CNNs are hierarchical. However, current implementation of NTI only operates on non-overlapping sub-trees while CNNs can slide over the input to produce higher-level representations. 
NTI is flexible in selecting the node function and the attention mechanism. Like CNN, the computation in the same tree-depth can be parallelized effectively; and therefore NTI is scalable and suitable for large-scale sequence processing.
Note that NTI can be seen as a generalization of LSTM. If we construct left-branching trees in a bottom-up fashion, the model acts just like sequential LSTM. 
Different branching factors for the underlying tree structure have yet to be explored. NTI can be extended so it learns to select and compose dynamic number of nodes for efficiency, essentially discovering intrinsic hierarchical structure in the input.  

\section*{Acknowledgments}

We would like to thank the anonymous reviewers for
their insightful comments and suggestions.
This work was supported in part by the grant HL125089 from the National Institutes of Health (NIH). Any opinions, findings and conclusions or recommendations expressed in this material are those of the authors and do not necessarily 
reflect those of the sponsor.

\bibliographystyle{eacl2017.bst}
\bibliography{eacl2017.bib}

\end{document}